\documentclass[10pt,twocolumn,letterpaper]{article}

\usepackage{iccv}              

\definecolor{iccvblue}{rgb}{0.21,0.49,0.74}
\usepackage[pagebackref,breaklinks,colorlinks,allcolors=iccvblue]{hyperref}

\usepackage{multirow}
\usepackage{pifont}
\usepackage{makecell}
\usepackage{xcolor}
\usepackage{algorithm}
\usepackage{algorithmic}
\usepackage{stfloats}
\definecolor{darkgreen}{rgb}{0.17,0.56,0.36}
\newcommand{\proposed}{RPEP}

\usepackage{soul} 
\usepackage{color} 

\def\paperID{2} 
\def\confName{ICCV}
\def\confYear{2025}

\title{Leveraging RGB Images for Pre-Training of \\Event-Based Hand Pose Estimation}

\author{Ruicong~Liu \textsuperscript{\rm 1, 2}\qquad~Takehiko Ohkawa\textsuperscript{\rm 1}\qquad~Tze Ho Elden Tse\textsuperscript{\rm 2}\qquad~Mingfang Zhang\textsuperscript{\rm 1}\\ Angela Yao\textsuperscript{\rm 2}\qquad~Yoichi Sato\textsuperscript{\rm 1}\\
		{\textsuperscript{\rm 1} The University of Tokyo, Japan}  
		\qquad~{\textsuperscript{\rm 2} National University of Singapore, Singapore} \\
	}

\begin{document}
\maketitle
\begin{abstract}
This paper presents \proposed, the first pre-training method for event-based 3D hand pose estimation using labeled RGB images and unpaired, unlabeled event data.
Event data offer significant benefits such as high temporal resolution and low latency, but their application to hand pose estimation is still limited by the scarcity of labeled training data. 
To address this, we repurpose real RGB datasets to train event-based estimators.
This is done by constructing pseudo-event-RGB pairs, where event data is generated and aligned with the ground-truth poses of RGB images.
Unfortunately, existing pseudo-event generation techniques assume stationary objects, thus struggling to handle non-stationary, dynamically moving hands.
To overcome this, \proposed~introduces a novel generation strategy that decomposes hand movements into smaller, step-by-step motions.
This decomposition allows our method to capture temporal changes in articulation, constructing more realistic event data for a moving hand.
Additionally, \proposed~imposes a motion reversal constraint, regularizing event generation using reversed motion.
Extensive experiments show that our pre-trained model significantly outperforms state-of-the-art methods on real event data, achieving up to 24\% improvement on EvRealHands. Moreover, it delivers strong performance with minimal labeled samples for fine-tuning, making it well-suited for practical deployment. 
\end{abstract}

\section{Introduction}
\label{sec:intro}

Capturing 3D hands from RGB images has been studied extensively~\cite{H:pavlovic1997visual,H:rautaray2015vision,H:von2001bare,H:holl2018efficient,H:piumsomboon2013user,H:ohkawa2023survey,H:fan2024egohand,H:xie2024learning}, but it remains vulnerable under challenging conditions such as wide brightness variation, complex lighting, and fast hand motion \cite{EV:jiang2024complementing, H:park20243d}.
Event cameras~\cite{EV:lichtsteiner2008128} are an alternative to RGB cameras that asynchronously capture per-pixel intensity changes.  
They can record high dynamic range images with an ultra-high frame rate (up to 1 \textmu s latency) \cite{H:rudnev2021eventhands, H:jiang2024evhandpose,EV:zheng2024eventdance,EV:long2024spike,EV:shi2024polarity,EV:shi2023identifying}.

Despite their advantages, datasets with real event data for 3D hand pose estimation are scarce due to the difficulty of annotation. 
Unlike RGB images, event data lacks texture information, making it difficult to annotate accurate 3D hand poses.
The EvRealHand dataset \cite{H:jiang2024evhandpose} is the first and only large-scale dataset to provide annotated real event data of hands. 
Although it offers event streams with 3D hand annotations, it heavily relies on a complex rig equipped with synchronized multi-camera RGB and event sensors. 
Such reliance on a studio-style capture rig limits its diversity and authenticity.

\begin{figure}[t]
\centering
\includegraphics[width=\linewidth]{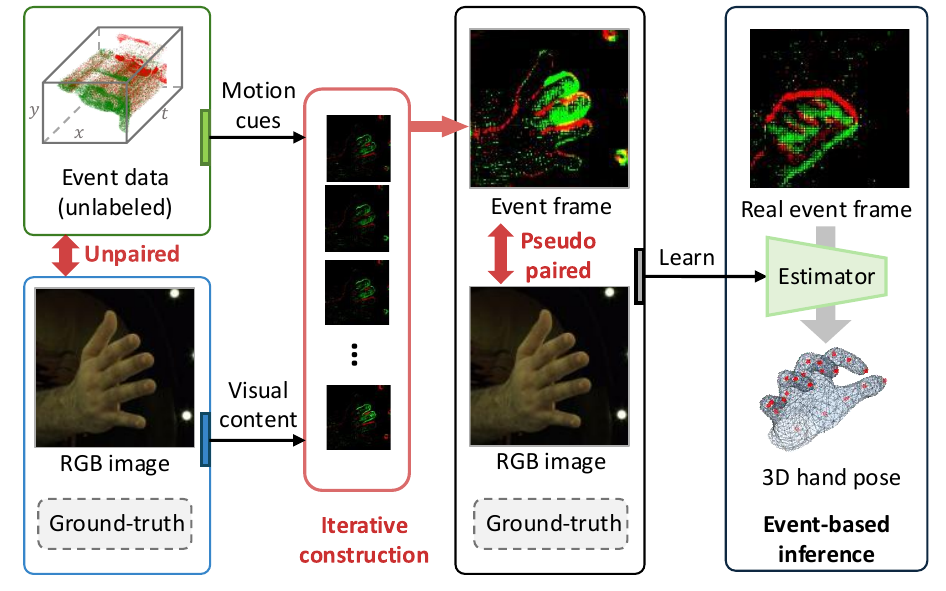}
\caption{We propose a pre-training method for event-based hand pose estimation using labeled RGB images and unpaired, unlabeled event data. At the core of our approach is an iterative construction module that generates a pseudo-event frame for each input RGB image, forming pseudo RGB-event pairs that reflect dynamic hand movements.
}
\label{fig:teaser}
\end{figure}

\begin{figure*}[t]
\centering
\includegraphics[width=\linewidth]{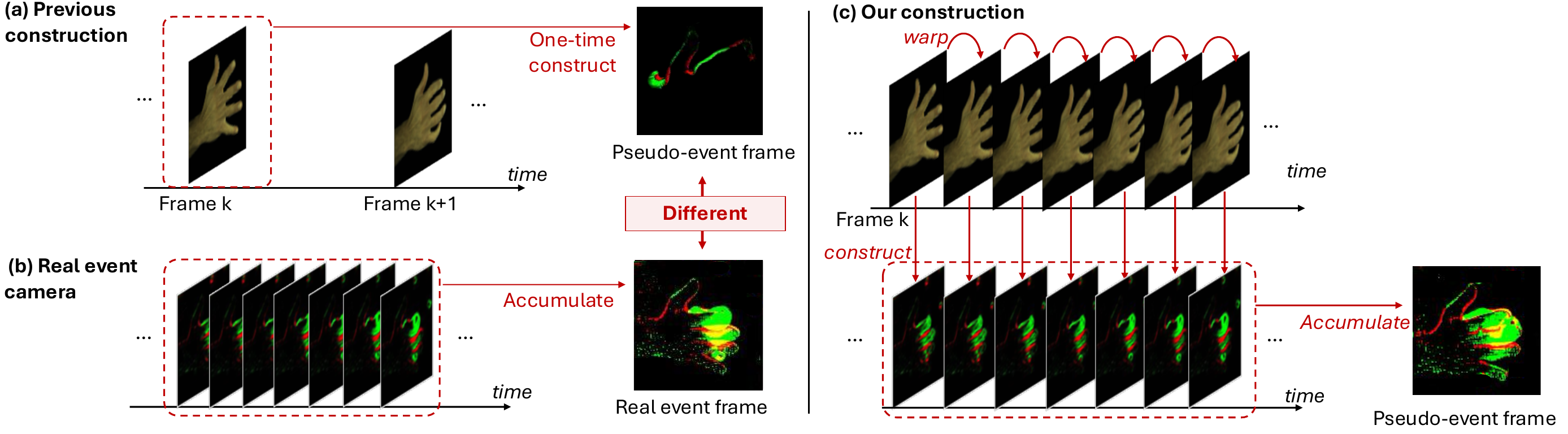}
\caption{Comparison of (a) RGB-event pair construction of previous methods \cite{EV:gehrig2020eklt,D:messikommer2022bridging}, (b) real event camera's capture process, and (c) our construction method for hands.
}
\label{fig:dynamic}
\end{figure*}

The lack of evnt datasets motivates our approach called \textbf{\proposed}~(\cref{fig:teaser}): leveraging \textbf{R}GB images for \textbf{P}re-training of \textbf{E}vent-based hand \textbf{P}ose estimation, helping to reduce the dependence on event annotations. 
To learn event-based models from RGB images, previous studies~\cite{EV:sun2022ess,D:messikommer2022bridging} construct pseudo-paired RGB–event data.
In this process, the generated pseudo-event frames are aligned with the ground-truth pose of the RGB image.
This alignment allows using the ground-truth hand pose annotations of the RGB images to train event-based estimators.
However, existing algorithms for constructing RGB–event pairs~\cite{EV:gehrig2020eklt,D:messikommer2022bridging} are not well suited for hand data, due to their poor reconstruction quality.
As shown in \cref{fig:dynamic} (a), previous methods generate event frames that are sparsely distributed along image edges.
In contrast, the real event frame in \cref{fig:dynamic} (b) exhibits a denser distribution in the finger regions, where finger articulations naturally trigger events.
Such discrepancy in event distribution brings a huge domain gap between real and constructed event data, which hinders the learning process and degrades the model's generalization to real environment.

Such poor construction quality stems from the stationary assumption underlying existing algorithms. As shown in \cref{fig:dynamic} (a), prior methods \cite{EV:gehrig2020eklt,D:messikommer2022bridging} generate pseudo-events only once from a single RGB frame.
This one-time construction implicitly assumes that the hand undergoes only rigid pose changes (\ie, translation and rotation) between frame $k$ and frame $k+1$.
It thereby ignores non-rigid articulations, which will induce additional events within this interval.
Consequently, the constructed events appear only along static image edges.
In contrast, due to the high temporal resolution of event cameras, real sensors capture much denser articulations that occur from frame $k$ to $k+1$, as illustrated in \cref{fig:dynamic} (b). These events are distributed across articulation regions such as moving fingers and the palm, covering all areas of motion. 

To construct more realistic pseudo-event data, we reformulate the original one-time construction into a process that simulates the real event accumulation process, as shown in \cref{fig:dynamic} (c).
From a single frame $k$, our method generates multiple intermediate RGB frames by warping, simulating the image changes caused by articulations.
For each RGB frame, we then perform a one-time construction \cite{EV:gehrig2020eklt,D:messikommer2022bridging} to generate corresponding pseudo-event frame.
Over time, the final pseudo-event frame is generated by accumulating all previous pseudo-event frames.
Unlike prior methods \cite{EV:gehrig2020eklt,D:messikommer2022bridging}, our construction process allows for articulations of the hand, thereby facilitating a more authentic simulation of the event accumulation.

To achieve the above process, we implement an iterative construction module, which separates the event accumulation into $T$ iterations.
At each iteration, this module generates an optical flow map to both 1) warp the RGB image and 2) construct pseudo-event frame.
As the process progresses, the image evolves, leading to the construction of event data that captures dynamic articulations.
To estimate the optical flow map, we use the RGB image and an unpaired, unlabeled event frame as input.
In detail, we extract 1) the \textit{visual appearance feature} of the RGB input and 2) the \textit{motion priors}, such as direction and trajectory, from the event input.
The motion priors provide necessary movement information, thereby enabling the static RGB hand to ``articulate".
The appearance feature and motion priors are then fed into a decoder to estimate the flow map.

Additionally, \proposed~imposes a novel motion reversal constraint, ensuring the the semantic correctness of the motion priors.
In other words, it ensures that the motion prior truly represent information such as physical moving direction and trajectory.
Starting from a constructed pseudo-event frame, we create a reversal frame with a completely opposite motion direction and trajectory.
Our method then maximizes the difference between the two motion priors from pseudo-event and reversal frames.
This constraint ensures the consistency between the extracted motion priors and the physical motion dynamics.

We evaluate our method across multiple challenging scenarios (\eg, flash and strong light) using the EvRealHands dataset \cite{H:jiang2024evhandpose} as the evaluation target.
Extensive experimental results demonstrate the superior performance of our method compared to existing transfer learning \cite{N:gehrig2020video,N:zhu2017unpaired,D:tzeng2017adversarial,D:messikommer2022bridging} and pre-training \cite{N:chen2020simple} methods.
In comparison to these methods, \proposed~exhibits a relative improvement rate of 24\% in normal scenes, 20\% in strong light scenes, and 14\% in flash light scenes.

\begin{figure*}[t]
\begin{center}
    \includegraphics[width=.95\linewidth]{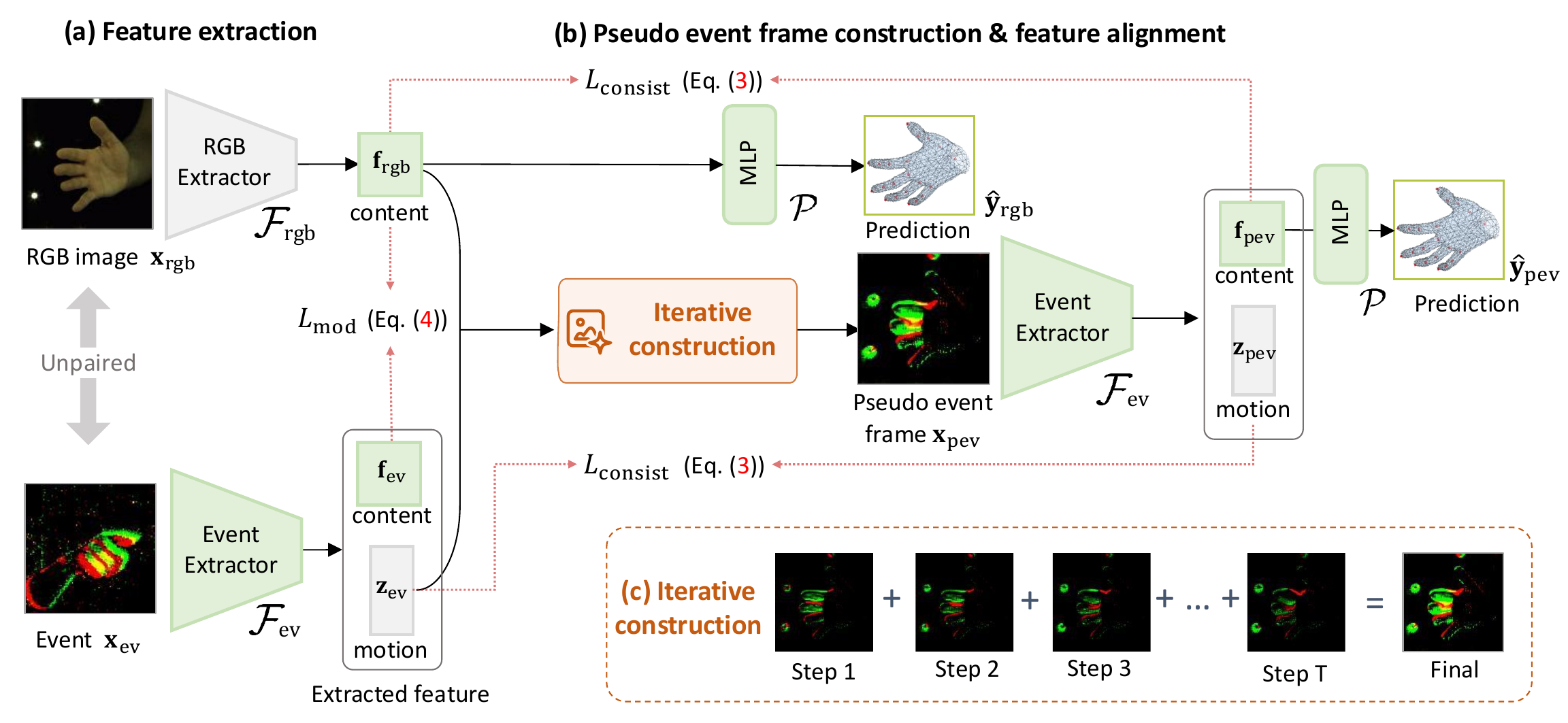}
\end{center}
\vspace{-5mm}
\caption{Overview of the proposed method, where labeled RGB images and unpaired, unlabeled event frames serve as inputs for training. (a) Feature extractors, $\mathcal{F}_{\text{rgb}}$ and $\mathcal{F}_{\text{ev}}$, extract features from their respective inputs. (b) Our iterative construction module constructs pseudo-event frames from these features, which are then fed into $\mathcal{F}_{\text{ev}}$ for feature alignment. The features $\mathbf{f}_{\text{rgb}}$, $\mathbf{f}_{\text{ev}}$, and $\mathbf{f}_{\text{pev}}$ are all aligned to the same feature space. (c) Illustration of the iterative construction process. Event frames from all iterations are accumulated together to get the final pseudo-event frame.
}
\vspace{-4mm}
\label{fig:overview}
\end{figure*}

\section{Proposed Method}\label{sec:method}
We introduce \proposed, a novel pre-training method for event-based hand pose estimation leveraging RGB images.
Our goal is to learn an event-based hand pose estimator, $\mathcal{H}=\mathcal{F}_{\text{ev}}\,\circ\,\mathcal{P}$, which comprises a feature extractor $\mathcal{F}_{\text{ev}}$ and a multi-layer perceptron (MLP) $\mathcal{P}$. 

As illustrated by \cref{fig:overview} (a), our method employs two feature extractors: an RGB extractor $\mathcal{F}_{\text{rgb}}$ and an event extractor $\mathcal{F}_{\text{ev}}$.
They process unpaired inputs, an RGB image $\mathbf{x}_{\text{rgb}} \in \mathbb{R}^{H\times W \times 3}$ and event frame $\mathbf{x}_{\text{ev}}\in \mathbb{R}^{H\times W \times 2}$ (event histogram \cite{EV:maqueda2018event}), respectively.
We design $\mathcal{F}_{\text{ev}}$ to capture the following two features from $\mathbf{x}_{\text{ev}}$.
\textbf{1) The appearance feature $\mathbf{f}_{\text{ev}}$ represents the hand pose, 2) and the motion prioirs $\mathbf{z}_{\text{ev}}$ represents the moving direction and trace.}
The RGB extractor $\mathcal{F}_{\text{rgb}}$ solely extracts the appearance feature $\mathbf{f}_{\text{rgb}}$ from $\mathbf{x}_{\text{rgb}}$, since the hand image is stationary and contains no motion information.
The $\mathbf{x}_{\text{rgb}}$ and $\mathbf{z}_{\text{ev}}$ are combined and fed into our iterative construction module to construct the pseudo-event frame $\mathbf{x}_{\text{pev}}$.

\noindent\textbf{Feature alignment.} 
To enable effective knowledge transfer and representation sharing across modalities, the MLP $\mathcal{P}$ is shared between both RGB and event. This encourages the extracted features $\mathbf{f}_{\text{ev}}$ and $\mathbf{f}_{\text{rgb}}$ to be aligned in the same latent space.
We further employ adversarial learning to explicitly align $\mathbf{f}_{\text{ev}}$ and $\mathbf{f}_{\text{rgb}}$.
In addition, we align features between the original input data and the constructed pseudo-event frames, \ie, $\mathbf{f}_{\text{rgb}}$ with $\mathbf{f}_{\text{pev}}$, and $\mathbf{z}_{\text{ev}}$ with $\mathbf{z}_{\text{pev}}$.

\noindent\textbf{Iterative construction.} To simulate to process of \cref{fig:dynamic} (c), we separate the time window $\Delta \tau$ into multiple iterations and develop an iterative construction module.
In each iteration $t$, we use a decoder $\mathcal{G}$ to generate an optical flow map $\mathbf{\hat{v}}^{(t)}$.
Its input contains the appearance feature of the input RGB image and the motion prior of the input event frame, \ie, $\mathbf{\hat{v}}=\mathcal{G}(\mathbf{f}_{\text{rgb}}, \mathbf{z}_{\text{ev}})$.
The estimated flow map $\mathbf{\hat{v}}^{(t)}$ is used for two purposes: \textbf{1) generating a sub-pseudo event frame} $\mathbf{x}_{\text{pev}}^{(t)}$, and \textbf{2) warping the RGB image} to reflect the dynamic image changes caused by articulation.
After $T$ iterations, all event frames are accumulated to form the final pseudo-event frame $\mathbf{x}_{\text{pev}}$:
\begin{equation} \label{eq:accumulate}
\mathbf{x}_{\text{pev}}=\sum_{t=1}^T \mathbf{x}_{\text{pev}}^{(t)}= \lfloor \sum_{t=1}^T \nabla \mathbf{x}_{\text{rgb}}^{(t-1)} \cdot \mathbf{\hat{v}}^{(t)} \rfloor.
\end{equation}

\begin{figure}[t]
\begin{center}
    \includegraphics[width=.98\linewidth]{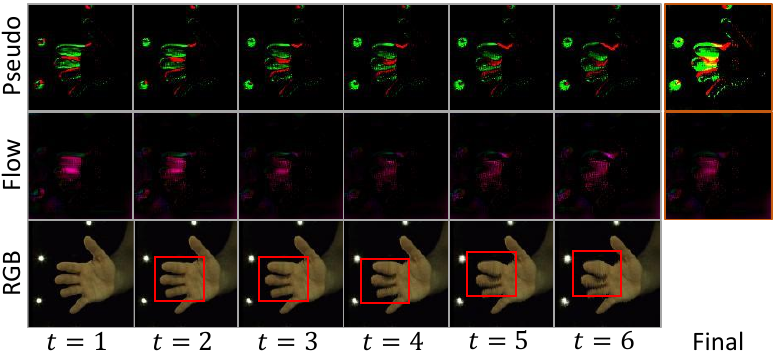}
\end{center}
\vspace{-5mm}
\caption{Constructed event frame, flow map, and RGB image from each iteration, with image changes caused by articulation highlighted in red rectangles.
}
\vspace{-5mm}
\label{fig:iteration}
\end{figure}

\noindent This way, our construction process allows
for variations in the RGB image, thereby facilitating a more
authentic simulation of the event accumulation process.
Empirically, we set $T=6$.
\cref{fig:iteration} displays the constructed event frame, flow map, and warped RGB image of each iteration.

\begin{table*}[t]
\caption{Comparison with state-of-the-art methods. The evaluation set EvRealHands \cite{H:jiang2024evhandpose} is separated according to collection scenarios: ``Normal", ``Strong light", and ``Flash". Within each scenario, samples are further divided into ``Scripted" and ``Unscripted" hand poses.} \label{tab:sota}
\vspace{-2mm}
\begin{center}
\small
\renewcommand\arraystretch{.9}
\setlength{\tabcolsep}{3.4mm}
\begin{tabular}{clccccccc}
    \toprule[1.2pt]
    
    &\multirow{2}{*}{Method}&\multirow{2}{*}{Metrics} & \multicolumn{2}{c}{Normal} & \multicolumn{2}{c}{Strong light} & \multicolumn{2}{c}{Flash} \\
    && & {Scripted} & {Unscripted} & {Scripted} & {Uncripted} & {Scripted} & {Uncripted} \\
    
    \hline \specialrule{0em}{1pt}{1pt}
    
    \multirow{4}{*}{\rotatebox{90}{Baseline}} & \multirow{2}{*}{w/o pre-train} & 3D-MPJPE & {27.98} & {49.12} & {31.95} & {52.94} & {36.20} & {51.53} \\
    && PA-MPJPE& {13.38} & {17.95} & {16.96} & {19.36} & {15.04} & {22.29}\\

    \cline{2-9} \specialrule{0em}{1pt}{1pt}
    
    &\multirow{2}{*}{Synthetic \cite{H:rudnev2021eventhands}}& 3D-MPJPE & {37.04} & {56.77} & {37.76} & {56.73} & {37.33} & {57.70} \\
    && PA-MPJPE& {18.78} & {20.09} & {20.47} & {20.60} & {17.42} & {25.76}\\

    \hline \specialrule{0em}{1pt}{1pt}
    
    \multirow{12}{*}{\rotatebox{90}{Main results}} &\multirow{2}{*}{SimCLR \cite{N:chen2020simple}}& 3D-MPJPE & 32.72 & 52.49 & 30.94 & 53.90 & 41.28 & {58.04} \\
    && PA-MPJPE& 13.92 & 17.14 & 16.41 & 17.88 & 15.40 & {22.95} \\
    \cline{2-9} \specialrule{0em}{1pt}{1pt}
    
    &\multirow{2}{*}{Vid2E \cite{N:gehrig2020video}}& 3D-MPJPE & 25.92 & 44.59 & 26.83 & 47.47 & 33.81 & {51.39}\\
    && PA-MPJPE& 13.32 & 16.89 & 15.17 & 18.94 & 14.72 & {20.80}\\
    \cline{2-9} \specialrule{0em}{1pt}{1pt}
    
    &\multirow{2}{*}{CycleGAN \cite{N:zhu2017unpaired}}& 3D-MPJPE & 24.32 & 46.78 & 28.57 & 50.65 & 31.27 & {51.03}\\
    && PA-MPJPE& 13.13 & 16.34 & 15.95 & 18.08 & 15.08 & {21.98}\\
    \cline{2-9} \specialrule{0em}{1pt}{1pt}
    
    &\multirow{2}{*}{ADDA \cite{D:tzeng2017adversarial}}& 3D-MPJPE & 25.79 & 45.80 & 31.06 & 51.02 & 34.25 & {49.17}\\
    && PA-MPJPE& 13.32 & 16.35 & 16.41 & 18.28 & 15.27 & {22.09}\\

    \cline{2-9} \specialrule{0em}{1pt}{1pt}
    
    &\multirow{2}{*}{RPG-EV \cite{D:messikommer2022bridging}}& 3D-MPJPE & 29.51 & 47.81 & 34.34 & 53.19 & 34.15 & {51.59}\\
    && PA-MPJPE& 15.51 & 16.89 & 18.05 & 18.58 & 16.56 & {22.61}\\

    \cline{2-9} \specialrule{0em}{1pt}{1pt}
    
    &\multirow{2}{*}{\proposed~(Ours)} & 3D-MPJPE & \textbf{21.26} & \textbf{39.91} & \textbf{26.08} & \textbf{44.28} & \textbf{30.97} & {\textbf{48.55}}\\
    && PA-MPJPE& \textbf{12.11} & \textbf{15.45} & \textbf{14.47} & \textbf{17.85} & \textbf{14.06} & {\textbf{20.11}}\\
    \bottomrule[1.2pt]
\end{tabular}
\vspace{-3mm}
\end{center}
\end{table*}

\noindent\textbf{Motion reversal constraint}
To ensure that the motion priors, $\mathbf{z}_{\text{ev}}$ and $\mathbf{z}_{\text{pev}}$, truly reflect actual physical dynamics, we introduce a motion reversal constraint.
Our method reverses the flow map $\mathbf{\hat{v}}^{(t)}$ across iterations from 1 to $T$, generating reverse events $\mathbf{x'}_{\text{pev}}^{(t)}$ as:
\begin{equation} \label{eq:reverse}
\mathbf{x'}_{\text{pev}}=\sum_{t=1}^{T}\mathbf{x'}_{\text{pev}}^{(t)}=\lfloor\sum_{t=1}^{T}\nabla \mathbf{x}_{\text{rgb}}^{(t+1)} \cdot -\hat{\mathbf{v}}^{(t)}\rfloor.
\end{equation}

\noindent Here, we reverse the direction of motion using $-\hat{\mathbf{v}}^{(t)}$ and invert the trace by employing $\mathbf{x'}_{\text{pev}}^{(t+1)}$ from the subsequent iteration $t+1$.

Given that the movement direction and trace of $\mathbf{x'}_{\text{pev}}$ are opposite to those of $\mathbf{x}_{\text{pev}}$, their motion features, $\mathbf{z'}_{\text{pev}}$ and $\mathbf{z}_{\text{pev}}$, should exhibit high divergence.
Therefore, we propose a divergence loss $\mathcal{L}_{\text{div}}$ to minimize their cosine similarity.

\section{Experiment}
In \cref{tab:sota}, we compare our method with state-of-the-art approaches. 
In all experiments, we first pre-trains an event-based pose estimator from labeled RGB data from InterHand2.6M \cite{H:moon2020interhand2} and unlabeled event data from EvRealHands\cite{H:jiang2024evhandpose}. 
Then, we fine-tune the pre-trained estimator using a few labeled samples from the EvRealHands.

The compared methods include SimCLR~\cite{N:chen2020simple}, an unsupervised pre-training approach; Vid2E~\cite{N:gehrig2020video} and CycleGAN~\cite{N:zhu2017unpaired}, which convert RGB videos into event representations; ADDA~\cite{D:tzeng2017adversarial}, a domain adaptation technique; and RPG-EV~\cite{D:messikommer2022bridging}, a transfer learning method that utilizes labeled RGB data along with unlabeled event data to train event-based networks.
For completeness, we also include two baselines: 1) fine-tuning a randomly initialized estimator without any pre-training, and 2) pre-training on a synthetic event dataset, EventHands~\cite{H:rudnev2021eventhands}.

\cref{tab:sota} presents the results after fine-tuning, showing that our method consistently outperforms all state-of-the-art approaches under various lighting conditions, achieving the lowest MPJPE error in every case.
Interestingly, the ``Synthetic” setup performs worse than even the ``w/o pre-train” baseline, suggesting the huge domain gap between synthetic and real event data.
Most other RGB-pre-training methods outperform the baselines, highlighting the benefit of using real RGB data for pre-training.
Notably, our method surpasses RPG-EV by a significant margin—achieving an 8mm lower error—demonstrating the effectiveness of our iterative pseudo-event construction in improving estimation performance.

\section{Conclusion}
In this paper, we introduce \proposed, a novel pre-training method for event-based 3D hand pose estimation leveraging RGB images. The core innovation of \proposed~is its iterative construction module, which generates pseudo-event frames that effectively accommodate dynamic hand motions. Additionally, our method incorporates a motion reversal constraint to refine the extracted motion priors, leading to enhanced construction results. Evaluation results demonstrate that \proposed~outperforms state-of-the-art techniques, achieving significant performance gains across a range of challenging scenarios. 
{
    \small
    \bibliographystyle{ieeenat_fullname}
    \bibliography{main}
}

\end{document}